# A gentle introduction to Quantum Natural Language Processing

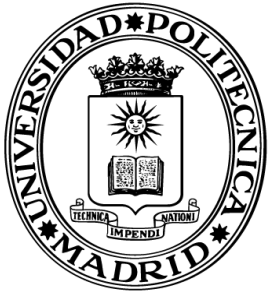

Shervin Le Du, UPM

A thesis submitted for the degree of :
Master in Quantum Computing Technology

Supervised by

Senaida Hernández Santana, PhD
Giannicola Scarpa, PhD

2021

*Dedication*

*I dedicate this thesis to my spouse, mother, family, teachers and friends who have thought me to achieve my dreams through patience and commitment.*

# Abstract

The main goal of this master's thesis is to introduce Quantum Natural Language Processing (QNLP) in a way understandable by both the NLP engineer and the quantum computing practitioner. QNLP is a recent application of quantum computing that aims at representing sentences' meaning as vectors encoded into quantum computers. To achieve this, the distributional meaning of words is extended by the compositional meaning of sentences (DisCoCat model) : the vectors representing words' meanings are composed through the syntactic structure of the sentence. This is done using an algorithm based on tensor products. We see that this algorithm is inefficient on classical computers but scales well using quantum circuits. After exposing the practical details of its implementation, we go through three use-cases.

First we show that the DisCoCat framework has been used to do sentence similarity. While its straightforward implementation using quantum nearest neighbor is theoretically possible, its full implementation requires quantum RAM, a technology not yet available. We therefore review a hybrid classical-quantum workflow that was proposed to overcome that technical limitation. It encodes the dataset into a single superposition state and the distances within the amplitudes of the superposition.

Second, we show how the measured output of quantum circuits have been successfully used to do binary sentence classification. We see that sentences are mapped into quantum circuits with the use of parameters for rotation gates ($R_x$, $R_z$) which can be optimized classically for a given task and dataset (question-answering in our case). While this is the first QNLP algorithm implemented on a quantum computer, our replication of the experiments leads us to raise some concerns as to the model's universality and scalability. We also see solutions have been proposed to offer better scalability at the price of increasing the size of circuits.

Finally, we see an extension that was proposed to encode hyperonomy (hierarchical word structure denoting supertype) using mixed states and ordering of positive operators. We see that when used within the DisCoCat framework, this model performs well at predicting sentence entailment

While this thesis was designed to be a review of the literature about QNLP, we also introduce original experiments on question-answering. After replicating previous experiments on hand-labelled sentences, we extend the words' mappings to add coverage for logical connectors. These play a central role in compositionality as they allow to combine sentences in a way that the truth value of the whole is predictable from ones of its parts. The model we introduce to account for those, while being simplistic, is a promising implementation.

# Table of Contents

# I. Introduction

## A. Historical Background

The idea of Natural Language Processing (NLP) goes back as far as the birth Computer Science. Alan Turing, who is considered the father of Computer Science, proposed in 1950 what is now called the Turing test.[55] This sets the goal for Artificial Intelligence to become indistinguishable of humans on tasks involving natural language understanding and generation.

In the meanwhile, the 50's and 60's saw the rise of formal linguistics. This branch aims at formalizing linguistic structures into a mathematical model. The most famous protagonist of this emerging field is likely Noam Chomsky. His best known contribution being to formal grammar with the elaboration of Context-Free Grammars (CFG).[15] CFGs are made of a vocabulary and production rules, and are designed to produce (and recognize) every grammatical sentences for a given set of rules.

The parallel rise of computers and formalization of grammar lead to an initial enthusiasm as to the the perspectives of NLP. At the time, the algorithms were always rule-based : linguistic experts would design by hand the vocabulary and the set of linguistic rules that would be implemented on a computer to complete a given task.[16] In the broader perspective of Artificial Intelligence (AI), the rule-based techniques can be linked to the so-called *symbolic* approach : the algorithms are designed to mimic the cognitive steps of a human doing the same task. This means the algorithm should read and produce human-readable resources (rules, databases etc.).[14]

In the geopolitical context of the cold-war, the main targeted application of NLP was machine translation from Russian to English. In 1954, a proof of concept experiment lead by Georgetown University and IBM allowed the translation of 49 carefully selected Russian sentences.[16] The authors claimed that machine translation could be solved within the end of the decade. However, in 1964, concerned by the lack of substantial progress, the US government required the Automatic Language Processing Advisory Committee (ALPAC) to lead an investigation on the state of the technology. In 1966, in its report, the ALPAC concludes that it is slower, less accurate and twice as expensive as human translation, and that "there is no immediate or predictable prospect of useful machine translation".[1] This report brought most researches on machine translation to an end and concluded the initial optimism for symbolic approaches in NLP.

In the 90's a change of paradigm was operating in the field of NLP. The computational power had increased considerably, and the number of available linguistic data as well (especially with the birth of internet). Concurrently, NLP shifted from being a rule-driven technology to a data-driven one.[16] Algorithms were designed to produce their own "rules" in order to fit as close as possible to the existing data. This marks the rise of Machine Learning era and the decline of symbolic AI. Indeed the rules produced by the algorithms are no longer built in some correspondence to cognitive models. Unsurprisingly, many of the early successes of this new era of statistical NLP are in the field of machine translation. Noticeably, IBM's alignment models were used to map each words to its translation inside a multilingual corpora and provided a way to train efficient statistical model of translation.[4] These models would remain the state-of-the art in machine translation for over two decades.

However, during the past decade a technological revolution has taken place in the field of algorithmics that has set a new baseline to the performance of AI. The exponential increase of the computational power that can be used in parallel, noticeably through the development of GPUs and cloud computing, has enabled the efficient implementation of what is known as neural networks. These are also data-driven, but the key difference is that their structure rely on an analogy to the structure of the brain at the neuronal scale. Neural networks stack layers of artificial neurons that interconnect to one another. The algorithm runs into two steps : feed-forward, where the network predicts the answer to a given problem, and back-propagation, where given the actual correct answer to the task, the network readjusts the weights of every connection inside the network to improve its future predictions.[13] As opposed to symbolic AI, this approach has been called *connectionist* because it relies on mimicking the low level structure of the brain instead of the high-level steps of cognition. Up to the date, neural network technologies remain the state-of-art of many major tasks in NLP.

The last decade has also known the development of a relatively new type of hardware : quantum computers ; and field : quantum computing. This development has lead to increased efforts and financing of these emerging technologies. The main motivation is to discover and implement algorithms that run more efficiently on these hardwares than on classical computers. The most famous example is Shor's algorithm, which enables finding the prime factors of a given number exponentially faster than any known classical algorithm would.[48] Designing quantum algorithms for NLP tasks is a sub-field of quantum computing now called quantum natural language processing (QNLP).
Around 2008, QNLP is being pioneered in a collective effort of three researchers of the University of Oxford : Mehrnoosh Sadrzadeh, Stephen Clark and Bob Coecke. Mehrnoosh Sarzadeh was working on grammar algebra : a mathematical formalism to account for grammar in natural languages. Stephan Clark was working on word embedding : a technique to consistently model words as elements of a vectorial space in order to encode fundamental aspects of their meanings. Bob Coecke was working on categorical quantum mechanics : a way to describe quantum mechanics in terms of processes that can be composed together. In Mehrnoosh et al. (2010) they lay the foundations of what will become the Distributional Compositional Category (DisCoCat) model : a model that combines embedded words along the grammatical structure of a sentence to encode its meaning.[34] The algorithm which is proposed by the authors to implement this model has an exponential speed-up when implemented on a quantum computer. With this model that encodes linguistic structure into quantum circuits, there is a claim for putting symbolic AI back on stage and thus to provide algorithms that no longer act like black boxes as neural networks do.

## B. Outline of the Thesis

In the following sections, we will first define the DisCoCat model in further details, then go through three NLP use-cases that can be formulated in the DisCoCat framework along with some experiments and proofs of concepts for each of them. We will discuss the advantages and drawbacks of the quantum algorithms as compared to their classical counterparts. Finally, we will discuss the open challenges and perspectives for the field of QNLP.

In this work we will also introduce a series of experiments of our own: we first reproduce previous experiments made on question answering and adapt them to include hand-labelled data, we then propose an implementation to account for sentences composed through *logical connectors*.[28] Logical connectors play a central role in both linguistics and mathematics as they enable combining propositions into new ones.

This work was conceived as a gentle introduction to the QNLP field. It is aimed at both the NLP engineer and the quantum computer research scientist who want to get a general overview of the basis and the current state of this growing field. We assume some familiarity with basic linear algebra concepts. To the reader unfamiliar with these, we recommend referring to Lay (2016).[22] To the one who wants to deepen his knowledge of quantum computing we recommend Nielsen & Chuang (2010)[6] and to the one unfamiliar with syntax theory, we recommend Chomsky and Lightfoot (2009).[5]

# II. DisCoCat Framework

## A.  Syntactic Structure & Lambek's Pregroup Algebra

Speech is always displayed linearly : in time for its spoken form, visually when it is written. Yet sentences have an underlying tree structure in which nouns complement verbs, adjectives complement nouns etc. Every word has a type and possibly dependents that complement its meaning. The underlying dependency structure of a sentence is referred to as *syntax tree*.[5] The most straightforward representation of a syntactic tree is by displaying the words in an acyclic graph, hierarchically from its root to its leaves. Another way is by recursively enclosing the dependents into brackets. This is the format we will retain here.

*Example:*

> "Time flies like an arrow."
> [Time] flies [like [[an] arrow]]

*Grammar* refers to the rules for combining words that are permitted in such a syntactic tree for a given language. A typical problem in formal linguistics is one of knowing if a given sentence has a valid structure for a given grammar. For instance "Time ants like an arrow" does not break down into any valid syntax tree because it has no word that can consistently be typed as a verb. On the other hand, one sentence can be derived into various valid syntax trees which induce different meanings. For instance "John saw the man on the mountain with a telescope" can be interpreted differently depending on whether you link "with a telescope" to "John", "a man" or "the mountain".

The task of looking for all valid syntactic trees of a given sentence is referred to as *parsing*. In his work, Lambek provides a parsing method for natural languages which formally corresponds to the contraction operation in a pregroup algebra.[21] A pregroup algebra is a partially ordered algebra $(A, 1, \cdot, \_^l, \_^r, \leq)$ such that $(A, 1, \cdot)$ is a monoid satisfying :

$$x^l \cdot x \leq 1 \quad x \cdot x^r \leq 1 \quad \text{(contraction)}$$
$$1 \leq x \cdot x^l \quad 1 \leq x^r \cdot x \quad \text{(expansion)}$$

In a pregroup grammar like Lambek's, each word has a type. This type is made of an atomic type optionally decorated to its left (resp. right) by invert types that cancel-out when concatenated appropriately to their left (resp. right). We will indicate such an inverts with a superscript to the left (resp. right):

*Example:*

> *"arrow": $^{-1}det . n$, once concatenated to its left with a det, the remaining atomic type is noun*
> *"like": $adv . n^{-1}$, once concatenated to its right with a noun, the remaining atomic type is adverb*

The fact that the invert cancels out to the left or right asymmetrically is the reason the grammar's algebra is a pregroup and not a group. The parsing of a sentence is successful when all of its elements cancel out and only the special symbol *S* is left.

*Example:*

> *"an" : det*
> *"time" : n*
> *"arrow" : $^{-1}det . n$*
> *"flies" : $^{-1}n . S . adv^{-1}$*
> *"like" : $adv . n^{-1}$*
>
> *[Time] flies [like [[an] arrow]] :*
>
> *[n] . $^{-1}n . S . adv^{-1}$ . [adv . $n^{-1}$ . [[det] . $^{-1}det . n$]]*
> *→ [n] . $^{-1}n . S . adv^{-1}$ . [adv . $n^{-1}$ . [n]]*
> *→ [n] . $^{-1}n . S . adv^{-1}$ . [adv]*
> *→ S*
>
> *We see here that the parsing is successful.*

# B. Word Embedding and Distributional Hypothesis

When it comes to Natural Language Processing (NLP), most algorithms for sequence modeling simply discard the syntactic structure from their inputs. Bag of words and naive Bayes implementations are based on limiting context words to a given length.[57] Recurrent Neural Networks offer the possibility to increase this length by order of magnitude,[13] especially when attention mechanism is implemented.[56] Yet none of the traditional techniques involve a grammar aware method. We can arguably affirm that discarding the underlying structure leads to increased uninterpretability of the algorithm's prediction.[37]

While the sentence syntax is not generally encoded into the algorithm's inputs, there have been extensive efforts in encoding words' meaning into it. "Word meaning" in itself can be understood in different ways. The most intuitive interpretation is the cognitive one,[59] that we experience as humans every time we speak : the meaning of a word is what links it to the object it refers to. Second, we can consider the lexical meaning.[51] It can be thought as the one we can find in a dictionary: definition, grammatical category, synonyms, antonyms, gender, number etc. Some aspects of the cognitive meaning can be derived from the linguistic one, as the latter is its linguistic modelization. In NLP, a widely accepted paradigm refers to word meaning within the distributional hypothesis.[46] The latter states that words which have similar meanings appear in similar contexts. The interest is shifted from describing words' meanings to comparing words' meanings among them. In the following sections of this work, this is the paradigm in which *distributional* and *word meaning* have to be understood.

With the distributional hypothesis in mind, a very standard NLP pre-processing step is of building a vectorial space in which to represent individual word meaning.[18] Vectorial spaces offer a structure to operate sums and subtraction between vectors, as well as a notion of orthogonality and distance between vectors.[22] With the correct mapping from words to vectors, we can ensure that the linear operations and properties from vector spaces approach meaningful lexicographic equivalents :

*Example:*

$$[cat] + [baby] \approx [kitten]$$
$$dist([cat], [jaguar]_{feline}) \lessgtr dist([cat], [car])$$
$$[jaguar]_{feline} \cdot [jaguar]_{car} \approx 0$$

This task of consistently mapping individual words into a vector space modelizing their meanings is referred to as *word-embedding*. A straightforward implementation comes directly from the distributional hypothesis[18, 60] : we chose a set of context words, for instance the 2000 most common words of a corpus. These words then form the basis of the vector space in the following manner : for every word of the corpus, we count the frequency at which it appears 'near' to each context word. The latter frequencies are the coordinates of that word along the basis vectors.

*Example:*

> *Let the corpus:*
>
> *"kings have crowns" , "kings eat bones"*
> *"queens have crown" , "queens eat fish"*
> *"cats sleep on beds", "cats eat fish on beds"*
> *"dogs eat bones" , "dogs sleep on beds"*
>
> *Let the basis :*
> *["crowns", "beds", "bones", "fish"]*
>
> *The other words are then represented as :*
> *kings = [0.5, 0, 0.5, 0]*
> *queens = [0.5, 0, 0, 0.5]*
> *cats = [0, 1, 0, 0.5]*
> *dogs = [0, 0.5, 0.5, 0]*
> *have = [1, 0, 0, 0]*
> *eat = [ 0, 0.25, 0.5, 0.5]*
> *sleep = [0, 1, 0, 0]*

With this vectorial representation of words based on the distributional hypothesis, we expect to capture some insights into their lexical features (word similarity, opposite words etc.) that can be leveraged by further NLP algorithms.[18] In other words, the meanings of the individual words are modeled during the word-embedding step. Yet generally no equivalent modelization of the data is used to account for the meaning arising from syntactic structure.

## C.  Towards Sentence Embedding and Quantum Advantage

Once provided with a way to model word meaning through word embedding, we can legitimately ask if an equivalent exists to capture sentences' meaning. We will refer to the latter as the meaning arising from composing the words' meanings within a syntactic structure. From now on, this is the paradigm in which *compositional* and *sentence meaning* will be referred to. We essentially want a mathematical representation for sentences that we can calculate from its words' embeddings (distributional meaning of words) and its underlying syntactic structure (compositional meaning of sentences).

In P. Smolensky (1990), the author suggests using a representation based on *tensor product* : Tensor Product Representation (TPR), to capture syntactic bindings across sentences.[49, 50] Tensor product is here understood as the Kronecker product, denoted by ⊗. Given two matrices A, B we define A ⊗ B to be the block matrix where every element of A is multiplied by B (Fig. 2.1). TPR is a structure where every unit of a sentence can be part of the representation of every constituent. It can be compositionally encoded from the numerical vectors representing the composing words, and hence offers the ideal framework to build-up a structure to embed not only words, but also sentences.

$$\mathbf{A} \otimes \mathbf{B} = \begin{bmatrix} a_{11}\mathbf{B} & \cdots & a_{1n}\mathbf{B} \\ \vdots & \ddots & \vdots \\ a_{m1}\mathbf{B} & \cdots & a_{mn}\mathbf{B} \end{bmatrix}$$

Fig. 2.1 : Definition of tensor product

Based on Lambek's pregroup algebra, in Mehrnoosh et al. (2010), the authors introduce the CSC algorithm, which develops another implementation of tensor based representation of the meaning of a sentence.[34] This algorithm can be implemented on a quantum computer. Let the linear map $\Sigma \langle ii |$ (Fig. 2.2), which is the sum over all the basis vectors of the space N. We call this linear map a *cap*. It formally describes the state of $\log_n$ EPR pairs (a generalization of quantum entanglement). If the basis is of size $n$ and is orthonormal, the cap can simply be thought of as a row vector of size $n^2$, filled of 0, and valued 1 at the positions ⟨ii|. The steps of the CSC algorithm are detailed in Fig. 2.3.

$$\sum_i \langle ii | := \underset{\mathcal{N} \quad \mathcal{N}}{\smile}$$

Fig. 2.2 (Zeng & Coecke, 2016) : Definition of a cap

1. Compute the tensor product $\overrightarrow{words} = \vec{w}_0 \otimes \ldots \otimes \vec{w}_k$ in the order that each word appears in the sentence.

2. Construct a linear map that represents the grammatical type reduction by "wiring up" the vectors with the appropriate caps. For example, given that the pregroup type reduction of a noun/transitive verb/noun sentence is:

$$n \cdot {}^{-1}n \cdot s \cdot n^{-1} \cdot n \leq 1 \cdot s \cdot 1 \leq s$$

this linear map is:

$$\sum_i \langle ii | \otimes \text{id} \otimes \sum_i \langle ii |$$

3. Compute the meaning of the sentence by applying the linear map to the $\overrightarrow{words}$ vector. This results in a vector in $\mathcal{S}$ which corresponds to the meaning of the sentence. In the above example the result corresponds to the diagram:

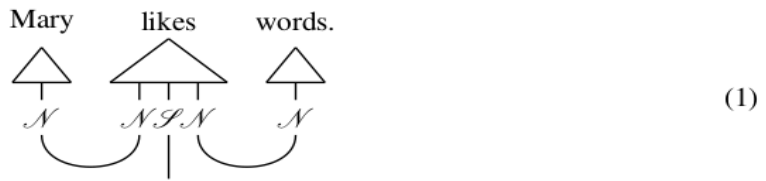 (1)

Fig. 2.3 (Zeng & Coecke, 2016) : the steps of the CSC algorithm. On the one hand, we calculate the tensor product of the words' vectors in the order they appear in the sentence. On the other hand, we parse the sentence and tensor the caps representing the bindings in the sentence. Finally, we compose the two tensor products.

On a classical machine the complexity of calculating full tensor products in a way that does not introduce further assumptions or inaccuracies is exponential.[60] So, as sentences increases in length, the CSC algorithm computational cost grows at an exponential rate. On a quantum device however, qubits intrinsically represent tensor products (Table 2.1),[60] and offer a way of implementing this algorithm that scales linearly with the size of the input.

For a sentence like "Mary likes words", considering a basis of size 2000 for the vector space N, storing the tonsorial product on a classical machine would require $2000^3$ bits : 8 x $10^9$ bits. In a quantum computer however, this can be stored in $\log_2(2000^3)$ qubits : 33 qubits. Considering 10'000 sentences, the classical storage required raises to 8 x $10^{13}$, whereas the number of qubits only increases to 47.[60]

|          | One Transitive Verb | 10k tr. verbs |
|----------|---------------------|---------------|
| Classical | $8 \times 10^9$ bits | $8 \times 10^{13}$ bits |
| Quantum  | 33 qubits           | 47 qubits     |

Table 2.1 (Zeng & Coecke, 2016) : Rough comparisons of the storage necessary for representing verbal phrases in quantum and classical frameworks.

The approach we have introduced consists in calculating the representation for a sentence from the vectorial representation of the words composing it along its syntactic structure. This approach is known by the name of distributional compositional categorical model of natural language meaning (DisCoCat). Quantum theory (as formulated by categorical quantum mechanics) has a mathematical structure very similar to DisCoCat's.[7, 8, 10, 37] This framework enables mappings from the pregroup grammar to quantum circuits. Such mappings will be referred to as *functors*.[29, 38] This means that NLP tasks formulated in the DisCoCat framework can be instantiated as quantum computations using functors.

# III. Use-case: Sentence Similarity

## A. Trying to leverage Quantum Nearest Neighbor Algorithm

A well-known classical algorithm used in Machine Learning is the nearest neighbor algorithm. Given a collection of vectors within a vectorial space, the task is to find the nearest neighbor(s) of a given vector. Nearest neighbor can be applied for clustering or label prediction (when there is a finite number of labels).[3] The naive approach has a complexity of O(MN), with M being the number of dimensions and N the number of vectors in the collection. In Wiebe et al. (2015), the authors have introduced a quantum variant for this algorithm that shows a quadratic speedup under certain conditions to its classical counterpart.[58]

However, the overhead from CSC algorithm would mitigate this speedup if applied naively. In Zeng & Coecke (2016), using some rewriting of the diagram resulting from CSC (Fig. 3.1) the authors provide a way to effectively leverage the quadratic speedup of the quantum version of nearest neighbors.[60]

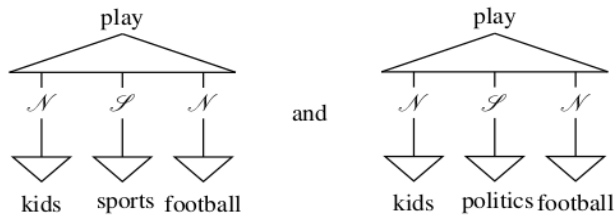

Fig. 3.1 (Zeng & Coecke, 2016) : instead of directly calculating the output state of the CSC algorithm, the diagram is rewritten into an equivalent diagram. The problem of looking for the nearest neighbor of "kids play football" among {"sports", "politics"} is brought to the one of looking for the nearest neighbor of "play" among {|kids⟩ ⊗ |sports⟩ ⊗ |football⟩, |kids⟩ ⊗ |politics⟩ ⊗ |football⟩}

Nevertheless, this algorithm will require further developing of quantum RAM to be taken full advantage of. Indeed once the tensorial representation for a sentence is instantiated on a quantum device, it should be stored in memory to be later retrieved in linear time during the calculations of nearest neighbors. Unfortunately, quantum RAM currently remains unrealized.

## B. A Hybrid Classical-Quantum Workflow for Sentence Similarity

In O'Riordan et al. (2020), the authors provide a hybrid classical-quantum workflow to implement sentence similarity on "noun-verb-noun" sentences without quantum RAM.[17, 39, 40]

First, distance is defined as the hamming distance. The dataset will be mapped into bitstrings, but the mapping must ensure that the hamming distances between the bitstrings reflect the underlying compositional and distributional relations between the sentences they represent. The authors provide pre-processing steps to ensure the aforementioned requirement is met (Fig. 3.2).

1. Tokenise the corpus and record position of occurrence in the text.

2. Tag tokens with the appropriate meaning space type (e.g. noun, verb, stop-word, etc.)

3. Separate tokens into noun and verb datasets.

4. Define basis tokens in each set as the $N_{\text{nouns}}$ and $N_{\text{verbs}}$ most frequently occurring tokens.

5. Map basis tokens in each respective space to a fully connected graph, with edge weights defined by the minimum distance between each other basis token.

6. Calculate the shortest Hamiltonian cycle for the above graph. The token order within the cycle is reflective of the tokens' separation within the text, and a measure of their similarity.

7. Map the basis tokens to binary strings, using a given encoding scheme.

8. Project composite tokens (i.e. non-basis tokens) onto the basis tokens set using representation cut-off distances for similarity, $W_{\text{nouns}}$ and $W_{\text{verbs}}$.

9. Form sentences by matching composite NOUN-VERB-NOUN tokens using relative distances and a NOUN-VERB distance cut-off, $W_{\text{nv}}$.

Fig. 3.2. (O'Riordan et al., 2020) : the classical procedure to consistently transform the dataset of sentences into a dataset of bitstrings. The key part of this procedure is to choose consistent mapping of the basis words into bitstrings, and this is done by calculating the shortest Hamiltonian cycle in the fully connected graph defined at step 5.

Following method from Trugenberger (2001), the authors then encode the dataset of bitstrings into a quantum register in the form of a unique superposition state representing the full dataset (Fig. 3.3).[54] This superposition state representing the full dataset is called the *test-pattern*. Following same method, we encode a *target vector* into a similar superposition state. The target vector is the sentence we want to know the closest vector to within the dataset.

Finally, the hamming distance between the target's state and each state representing a sentence in the dataset is encoded into the amplitudes of each state of the superposition. These are then retrieved through amplitude measurement, which is repeated multiple times in order to build a statistical distribution of the results.

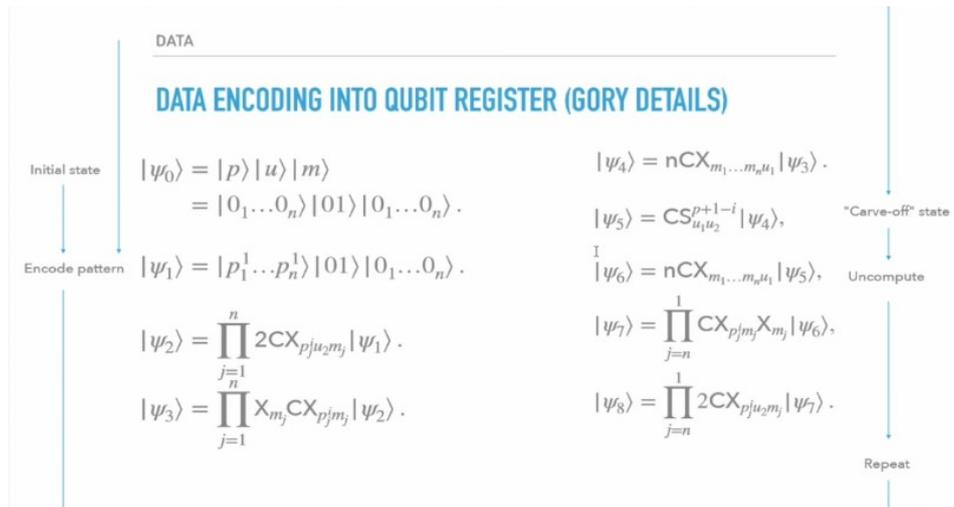

Fig. 3.3. (O'Riordan, 2019) : an iterative process is applied to the dataset of bitstrings in order to encode it into a quantum register in the form of a superposition state. This is done to overcome the limitation of not having access to quantum RAM.

## C. Proof of Concept

Using the workflow described in the previous section, the authors design a proof of concept experiment using the Intel Quantum Simulator to handle the quantum circuit workload.[17, 39, 40] Considering the basis vocabulary is made up of only 4 nouns, 4 verbs and 2 adverbs, the authors arbitrarily map each word to bitstrings, ensuring the hamming distances are consistent to intuitive distances between them (Table 3.1).

Because the authors only intend to design a proof of concept, instead of projecting the words onto the basis states using a formal word embedding method, they arbitrarily choose the components for each word in terms of the basis vectors. (Table 3.2)

They decide to restrict the dataset to two sentences : {J= "John rests inside", M= "Mary walks outside"}. By simply tensoring the words of each sentences, the latter are mapped to quantum states:

$$|J\rangle \rightarrow (½) ( |0\rangle \otimes (|10\rangle + |11\rangle) \otimes (|00\rangle + |10\rangle) )$$
$$= (½) ( |01100\rangle + |01000\rangle + |01110\rangle + |01010\rangle )$$
$$|M\rangle \rightarrow (½) ( |1\rangle \otimes (|00\rangle + |01\rangle) \otimes (|01\rangle + |11\rangle) )$$
$$= (½) ( |10011\rangle + |10111\rangle + |10001\rangle + |10101\rangle )$$

These states are then encoded into a superposition state representing the full dataset :

$$|m\rangle = (1/\sqrt{2}) (|J\rangle + |M\rangle)$$
$$= (1/\sqrt{2}) (|01100\rangle + |01000\rangle + |01110\rangle + |01010\rangle + |10011\rangle + |10111\rangle + |10001\rangle + |10101\rangle)$$

Finally, using the steps described in Fig. 3.3, they are able to build the probability distribution of the closest states (Fig. 3.4). Note that the dataset is made of two sentences corresponding to superposition states of 4 basis states each. On the other hand, the probability distribution only gives the hamming distance with the basis states. To infer the hamming distance with a superposition state one should sum-up the probabilities of the basis states that compose it.

Authors also run a more substantial experiment based on the corpus of sentences in *Alice in Wonderland*, increasing the nouns basis from 2 to 8 and increasing the basis states for sentences from 8 to 75. They use this new example to run the full algorithm, including the classical pre-processing part they eluded in the previous experiment. The only addition being the classical part of the algorithm, we won't detail this experiment any further here.

| Dataset | Token | Bin. Index |
|---|---|---|
| $n_s$ | adult | 00 |
| $n_s$ | child | 11 |
| $n_s$ | smith | 10 |
| $n_s$ | surgeon | 01 |
| $v$ | stand | 00 |
| $v$ | move | 01 |
| $v$ | sit | 11 |
| $v$ | sleep | 10 |
| $n_o$ | inside | 0 |
| $n_o$ | outside | 1 |

Table 3.1. (O'Riordan, 2020) : arbitrary mapping of the basis words to bit-strings. The mapping must ensure words we consider similar have closer bit-strings than words we consider as opposite. For instance, bitstrings for "adult" and "child" have the maximal hamming distance while ones for "sit" and "sleep" are closer to each other.

| Dataset | Token | State |
|---|---|---|
| $n_s$ | John | $(|00\rangle + |10\rangle)/\sqrt{2}$ |
| $n_s$ | Mary | $(|01\rangle + |11\rangle)/\sqrt{2}$ |
| $v$ | walk | $(|00\rangle + |01\rangle)/\sqrt{2}$ |
| $v$ | rest | $(|10\rangle + |11\rangle)/\sqrt{2}$ |
| $n_o$ | inside | $|0\rangle$ |
| $n_o$ | outside | $|1\rangle$ |

Table 3.2. (O'Riordan, 2020) : Using the bitstrings of the basis words, new words are described as superposition states of the basis words. For instance, "Walk" is modeled as a superposition of "stand" and "move."

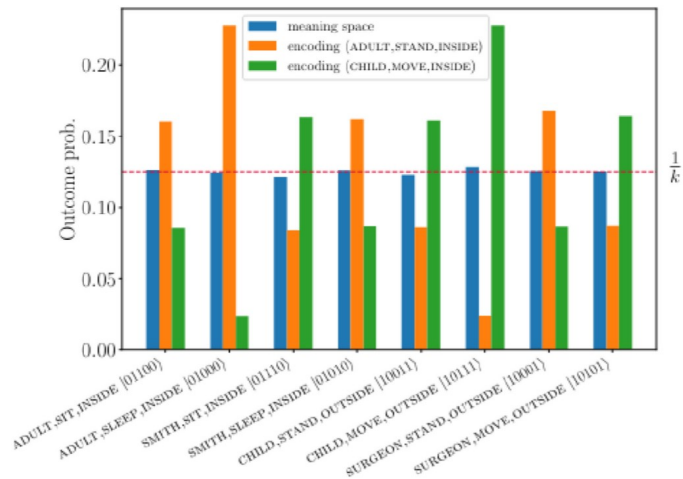

Fig. 3.4 (O'Riordan, 2019) : the distribution of the closest basis states to "adult(s) stand inside" (orange) and to "child(ren) move inside" (green) are each sampled 5 x 10⁴ times. For "adult(s) stand inside", we can see that the basis states composing |J⟩ have a higher overall probability than the ones composing |M⟩. Thus, for this example we can conclude the closest sentence within the database is "John rests inside". For the other example we can show its closest sentence within the database is "Mary walks outside"

# IV. Use-case: Question Answering

## A. Using DisCoPy to model Question-Answering

Question answering is the task of building systems that automatically answer questions formulated in natural language. If we restrain to yes/no questions, we can train a Machine Learning algorithm on a set of propositions with a binary label representing their truth value, and then use it to evaluate the truth values of sentences it has never seen before.

When using the DisCoCat framework to map syntactic tree of a sentence into a quantum circuit, this is typically done with the choice of certain parameters (phase of $R_x$, $R_z$, ...). The optimization of these parameters with regards to a given task is a classical optimization problem. One use-case of such a parameterized quantum computation in QNLP is question answering.

Using DisCoCat for question answering has been explored in much detail in K. Meichanetzidis et al (2020).[37] The authors have further provided a proof of concept experiment of their algorithm.[36, 37, 42, 43] They ran the experiment using the discopy python library, which is a library they had previously designed to simulate DisCoCat framework on classical computers.[11, 12, 41, 52] Subsequently, they also have it run on a real quantum device.[35, 37]

Let's first see how this task is handled by them on the quantum simulator. In the following we slightly reordered the steps followed by the authors in their notebook.[23, 36, 42, 43, 53] We took the liberty to do so for pedagogic reasons only.

1) First, define the atomic pregroup types (typically 's' and 'n').

2) Assign a number of qubits for each atomic pregroup type. In their work, 's' is always set to 0 qubits, because the qubit for the sentence is only intended to be measured, and hence is to be considered as a scalar.

3) Construct the finite vocabulary. For each word assign a pregroup type, using atomic pregroup types and invert types.

4) Every pregroup type will be mapped to a quantum circuit. Explicitly define the quantum circuits (Fig. 4.1, 4.2 and 4.3).

5) Explicit the mapping from each entry in the vocabulary to its corresponding quantum circuit.

6) Because the circuit is parameterized, define a function which takes a set of parameters as inputs, and outputs a functor from the grammar category to the category of quantum circuits. The functor outputs the quantum circuit that encodes the syntactic tree of any given grammatical sentence it is inputted with (Fig. 4.4).

7) Create another functor to evaluate the circuits.

8) Instantiate the dataset: a random set of grammatical syntax trees generated by the CFG. Ideally, we want these sentences to have their truth values labelled by hand, making sure that the truth values among the sentences are consistent. Alternatively, and this is what is done in by the authors in their proof of concept experiment, one can use the functor defined at steps 6 and 7 with arbitrary parameters to produce the labels (Fig. 4.5).

9) Use classical optimization techniques to choose the best parameters for the quantum circuits defined at step 4. In their work, the authors have used gradient descent.

10) Apply the functors from steps 6 and 7 on the test set, using parameters calculated at step 9. In their work, the testing score obtained is close to 100%.

Note that in further experiments, the authors switch to using a hand-labelled dataset (see following subsection for further discussion on this topic).[35, 37] Aside from that, in their work K. Meichanetzidis et al (2020) also adapt their experiment to have it run on a real quantum device instead of a simulator.[35, 37] By doing so, they run the first-ever QNLP experiment on a quantum computer. To do so, they compile the circuits to make it compatible with the processor's topology and to minimize the most noisy operations. The corpus they consider is called $K_{16}$ : a corpus of 16 sentences using 6 words. The experiment demonstrates a convergence of the cost function similar to the one observed on quantum simulators (Fig. 4.6).

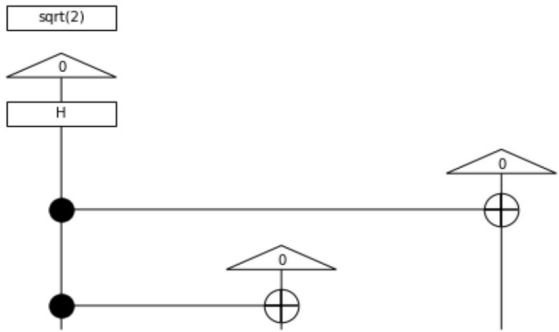

Fig. 4.1 (Oxford Quantum Group, 2020b): the circuit for the GHZ state has three outputs that can represent the three syntactic relations that has a relative pronoun like "who"[45]

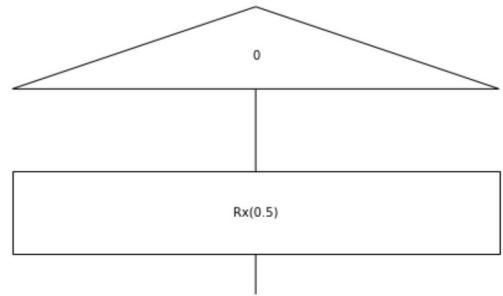

Fig. 4.2 (Le Du, 2021a): the circuit has one unique output that can represent the syntactic relation that has an intransitive verb like "is rich". Note that the circuit contains a $R_x$ gate whose phase is to be chosen.

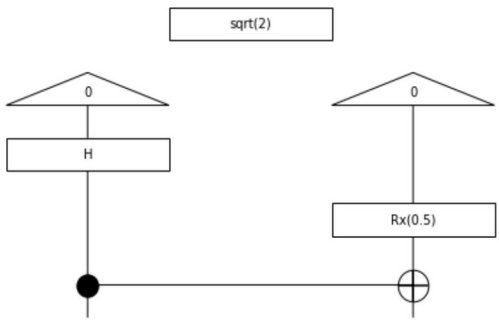

Fig. 4.3 (Le Du, 2021a): the circuit has two outputs that can represent the syntactic relations that has a transitive verb like "love". Note that the circuit contains a $R_x$ gate whose phase is to be chosen.

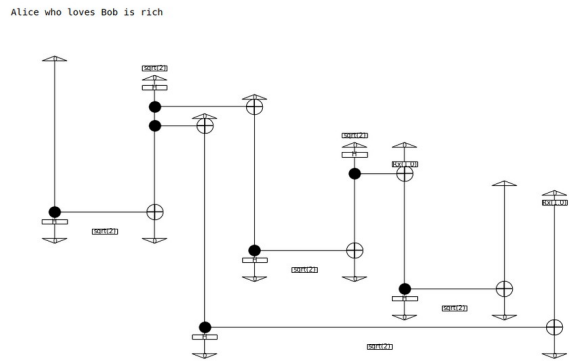

Fig 4.4 (Oxford Quantum Group, 2020a): quantum circuit encoding the syntactic tree of the sentence "Alice who loves Bob is rich"

```
True sentences:
Alice is rich.
Alice loves Bob.
Bob loves Alice.
Alice who is rich is rich.
Alice who loves Bob is rich.
Alice who is rich loves Bob.
Alice who loves Bob loves Bob.

False sentences:
Bob is rich.
Alice loves Alice.
Bob loves Bob.
Bob who is rich is rich.
Alice who loves Alice is rich.
Alice who is rich loves Alice.
Bob who loves Alice is rich.
Bob who loves Bob is rich.
Bob who is rich loves Alice.
Bob who is rich loves Bob.
Alice who loves Alice loves Alice.
Alice who loves Alice loves Bob.
Alice who loves Bob loves Alice.
```

Fig. 4.5 (Oxford Quantum Group, 2020b): arbitrarily choosing (0.5, 1.0) as parameters the authors assign a truth value to a set of 20 sentences generated using a Context Free Grammar.

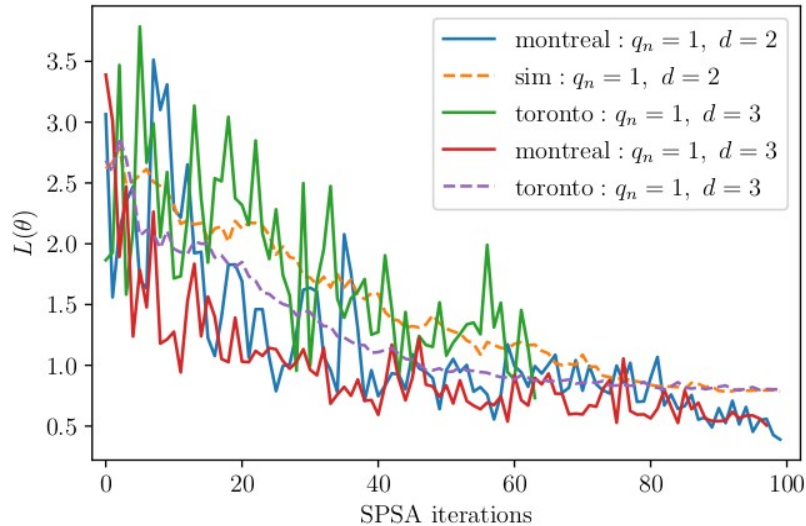

Fig. 4.6 (K. Meichanetzidis et al., 2020): evolution of the mean cost function ⟨L(θ)⟩ along the number of SPSA iterations of the optimization algorithm, against the corpus $K_{16}$ made of 16 sentences using a vocabulary of 6 words. Results are averaged over 20 realizations. Backends used are ibmq's quantum processors : ibmq toronto and ibmq montreal. The graph shows the cost function converges to 0 as the number of SPSA iterations increase.

## B. Our Original Experiments

### 1. Using Hand-Labelled Data

The first experiment we lead is directly inspired by the notebook described in previous section.[24] At step 8 of the algorithm, instead of instantiating labels from a functor using arbitrary parameters, we provide hand-labelled ones, ensuring logical consistency among the truth values (Fig. 4.7). This kind of arbitrary hand-labelled data is a less artificial case than the one provided in the original experiment. But by changing the dataset in this way, the gradient descent algorithm only obtains a training score of 77% and a testing score of 80%.

In the second experiment, we run the latter experiment introducing into the model one more intransitive verb : "is happy" and hence a new parameter.[25] In this case, we see the fitting score further falls to 62%. This leads to initial concerns as to the scalability of the model when the number of parameters increases (Table 4.1).

In their work, K. Meichanetzidis et al. (2020) have had some success in addressing the problem of scalability (Fig. 4.8).[37] Indeed, they manage to further improve the fitting score of their model by increasing the number of qubits assigned to the pregroup types (set to 1 in our experiments) and the depth of the circuit corresponding to each word (alternating layers of Hadamard gates and layers of controlled-Z rotation gates). While this is a promising workaround for the problem of the scalability of the model, it is only achieved at the cost of scaling-up the underlying circuits. This is problematic as it introduces further technical challenges because the size of quantum computers is currently limited.

|  | No hand-labelled data (50% - 50% train/test split) | Hand-labelled data (50% - 50% train/test split) | Hand-labelled data + extra parameter (50% - 50% train/test split) |
|---|---|---|---|
| Train score | 99.97% | 77% | 71.55% |
| Test score | 99.96% | 80% | 51.69% |
| Total fitting score (ignoring train/test split) | 99.97% | 78.61% | 62.1% |

Table 4.1: comparison of the scores of the original experiment with our first two experiments. The same dataset of 20 sentences is used in each experiment.

```
Alice is rich    1
Bob is rich      1
Alice loves Alice        0
Alice loves Bob 0
Bob loves Alice 1
Bob loves Bob    1
Alice who is rich is rich        1
Bob who is rich is rich 1
Alice who loves Alice is rich    0
Alice who loves Bob is rich      0
Alice who is rich loves Alice    0
Alice who is rich loves Bob      0
Bob who loves Alice is rich      1
Bob who loves Bob is rich        1
Bob who is rich loves Alice      1
Bob who is rich loves Bob        1
Alice who loves Alice loves Alice        0
Alice who loves Alice loves Bob 0
Alice who loves Bob loves Alice 0
Alice who loves Bob loves Bob    0
```

Fig. 4.7 (Le Du, 2021b) : we manually assign a truth value to the same set of 20 sentences as in Fig. 4.5

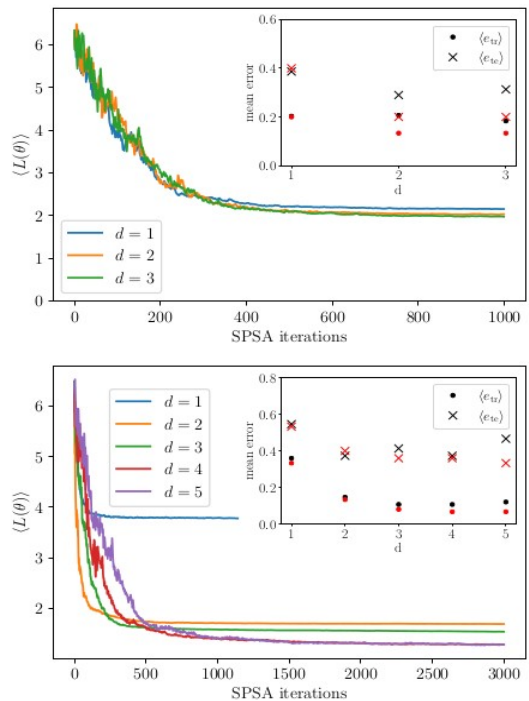

Fig. 4.8 (K. Meichanetzidis et al., 2020): evolution of the mean cost function ⟨L(θ)⟩ along the number of SPSA iterations of the optimization algorithm, against the corpus K$_{30}$ made of 30 sentences using a vocabulary of 7 words. Results are averaged over 20 realizations. Python is used as backend. Top graph shows the evolution when assigning 1 qubit for pregroup type $n$. The bottom one, when assigning 2 qubits for it. We see that the loss function decreases when adding more qubits to their representations (except for d=1). Moreover, we see that for a fixed number of qubits to represent $n$, the loss function decreases when increasing the depth $d$ of the circuit corresponding to each word.

## 2. Addition of Logical Connectors

Next, we run further experiments to model a different type of word : logical connectors.[26, 27] Under the DisCoCat framework the implementation of logical operators are of a great interest : they allow to compose recursively new sentences. In the context of question answering they have a second interest : they allow us to build sentences whose truth value can be logically deduced from its components. For instance, if "Alice loves Bob" is False, we expect "Alice does not love Bob" to be True.

The three logical connectors on which we focus in our experiments are : negation (not), conjunction (and), and disjunction (or). We will here propose an implementation for them and run proofs of concept experiments on this extended model. Note that we will limit to conjunction and disjunction of verbal phrases only, this is to say cases where there is one common subject for all the verbal phrases (i.e. "dogs chase cats and don't purr").

First, we change the assignment at step 1. Indeed, because '$s$' will be used as an input for logical connectors, '$s$' type has to be represented as a qubit instead of a scalar. We then give to the negation word the type $^{-1}n \cdot n$, and to conjunction and disjunction the type $^{-1}s \cdot n$. This enables a CFG to compose syntactic trees that have logical connectors in it (Fig. 4.9).

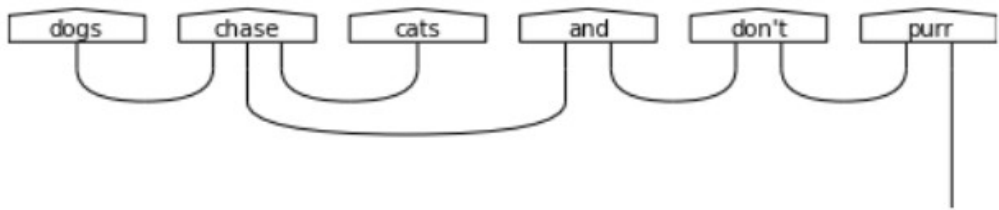

Fig. 4.9 (Le Du, 2021e) : after attributing appropriate types to logical connectors, we enable the Context Free Grammar to account for them.

Because the verbs now have the possibility to link to a logical connector, we need the circuits that represent them to have an extra output. Intransitive verbs are mapped to the circuit previously used for transitive verbs, and transitive verbs are mapped to the circuit for the GHZ state.

Because all of our logical operators have two relations (Fig. 4.9), we map them to the circuit we now use for intransitive verbs. Note that we introduce a new parameter for each logical connector. We therefore expect lower results than in previous experiment. Once each elementary circuit is defined, we can define the functor that maps any grammatical syntactic tree to its corresponding quantum circuit (Fig. 4.10). We also define the functor which will be used to evaluate the circuits.

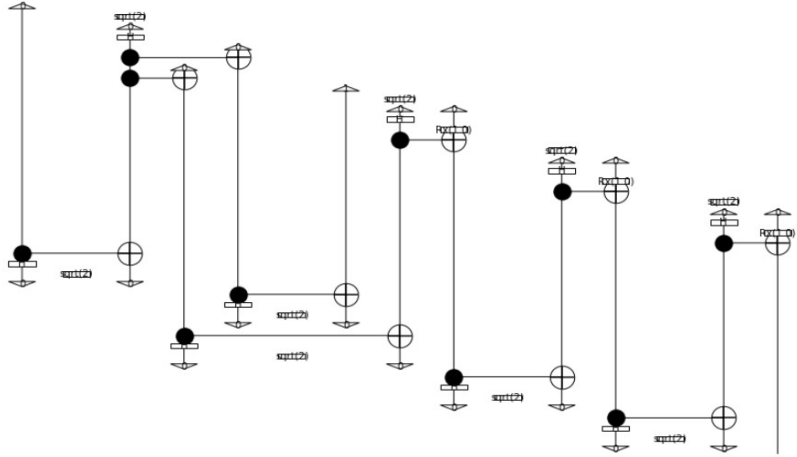

Fig. 4.10 (Le Du, 2021e) : the quantum circuit that encodes the syntactic tree of the grammatical sentence "dogs chase cats and don't purr". This quantum circuit is the output of the functor we have constructed to map syntactic trees to the quantum circuits encoding them.

The output of the circuit's evaluation through the second functor is a qubit. We use it to make the prediction as to the truth value of the corresponding sentence, by looking at whether the qubit is closer to $|0\rangle$ or $|1\rangle$. Because this experiment is meant to remain a proof of concept, we did not implement a proper optimization algorithm, and ran instead a random exploration of the space of the parameters. We do so because calculating gradients of the loss function with the jax library turned out to be technically harder to do with the aforementioned definition of the prediction function (see the Conclusions for further discussion on the issue). Despite this limitation of our approach, we find that when implementing only one logical door (negation), the total fitting score with the best parameters found is of 70%, and expectedly falls to 62% when implementing the three of them together (Table 4.2).

|  | One logical connector (50% - 50% train/test split) | Three logical connector (90% - 10% train/test split) |
|---|---|---|
| Train score | 50% | 58.88% |
| Test score | 90% | 70% |
| Total fitting score (ignoring train/test split) | **70%** | **62.62%** |

Table 4.2 : comparison of the scores implementing one vs three logical operators. A dataset of 20 sentences was used in the first case, and of 100 sentences in the second.

# 3. Discussion

As compared to classical algorithms for question answering, one can appreciate the interpretability of the function underlying the prediction. Indeed, the shape of the quantum circuit is directly inspired by the shape of the syntactic structure of the sentence it represents.

On the other hand, a legitimate concern can be addressed as to whether this framework corresponds to a universal model. For a chosen set of parameters, the algorithm gives a prediction of the truth value (0 or 1) of every example in the dataset. If the dataset is of size N, the output of the algorithm will be an element of $\{0, 1\}^N$. But given that the correct assignment of truth values is arbitrary among $\{0, 1\}^N$, we should attach a particular importance to proof or disproof the universality of the model, which could be defined as its ability to output any arbitrary element of $\{0, 1\}^N$.

To our knowledge, no such proof of universality exists. Nevertheless, we have seen that methods have been proposed to address the problem of scalability by increasing the size of the circuit.[37] While this is promising for the modeling properties of the aforementioned framework, this is done at the cost of increasing the hardware requirements. This is a problem at a time where quantum computers have limits to the number of qubits and gates they can implement.

# V. Use-case: Compositional Hyperonomy and Entailment

## A. Encoding Hyperonomy and Entailment into Positive Operators

Encoding semantically relevant features into the input of NLP Algorithms is an important pre-processing step. We have seen it in the example of word embedding and through the efforts put into providing a framework which encodes the syntactic structure of sentences. Other semantical features we haven't seen yet include *hyperonomy* and *entailement*. Hyperonomy refers to the relation between two words in which one is the supertype of the other (e.g. 'cat' → 'felin', 'felin' → 'animal'). For a given object, hyperonym refers to its supertype, while hyponym refers to its subtype. Entailment on the other hand refers to two sentences for which one cannot be true without the other one being true as well (e.g. 'Garfield is a lazy cat.' entails 'There are lazy animals').

Up until now, the settings we have considered had each word encoded into a deterministic sequence of qubits. This is to say each word was encoded into a *pure state*. Nevertheless, we can design a setting where instead of being deterministic, the sequence of qubits encoding a word is probabilistic. In this case the words are said to be encoded into *mixed states*. In quantum mechanics, uncertainty about the state of a system is expressed in what is called *density matrices*. A matrix $\rho$ is a density matrix if it meets following requirements :

- $\forall v \in V, \langle v|\rho|v\rangle \geq 0$ ,
- $\rho$ is self-adjoint, 1
- $\rho$ has trace 1.

Matrices satisfying only the two first conditions are called *positive operators*. The density matrix of a pure state $|v\rangle$ is given by its outer product : $|v\rangle\langle v|$. And the density operator of a mixed state is the weighted sum of the states composing it :

$$\rho = \sum_s p_s |\psi_s\rangle\langle\psi_s|$$

Mixed states and their matrix representation are leveraged in M. Lewis (2019) to provide an extension to the DisCoCat framework that also encodes hyperonomy and entailment within its structure.[30, 32, 33] To do so, the author considers the vectors representing elementary words (Fig. 5.1) to be the pure states, and conceptualizes a hyperonym as a mixture state of all its hyponyms. The author uses positive operators instead of density matrix because she implements another normalization process than having the trace equal to 1. The positive operator of a word corresponding to a pure state is constructed through the outer product with itself (Fig. 5.2), while the positive operator of a hyperonym is built by summing over the matrix representations of all its hyponyms (Fig. 5.3).[2]

|  | pug | goldfish | tabby |
|---|---|---|---|
| furry | 3 | 0 | 5 |
| domestic | 4 | 5 | 5 |
| working | 0 | 0 | 0 |
| aquatic | 0 | 6 | 0 |

Fig. 5.1 (M. Lewis, 2019) : vectors representing the pure states "pug", "goldfish" and "tabby" expressed using the basis of words {"furry", "domestic", "working", "aquatic"}

$$|cat\rangle \mapsto |cat\rangle\langle cat|$$

Fig. 5.2 (Bankova et al., 2019) : building the positive operator representing the elementary word "cat"

$$\begin{aligned}[\![pet]\!] &= \overline{pug} + \overline{goldfish} + \overline{tabby} \\ &= \vec{pug}\,\vec{pug}^\top + \vec{gfish}\,\vec{gfish}^\top + \vec{tabby}\,\vec{tabby}^\top \\ &= \begin{pmatrix} 34 & 37 & 0 & 0 \\ 37 & 66 & 0 & 30 \\ 0 & 0 & 0 & 0 \\ 0 & 30 & 0 & 36 \end{pmatrix}\end{aligned}$$

Fig. 5.3 (M. Lewis, 2019): construction of the positive operator of a hyperonym.

By building these operators in this way, the author ensures them to have two key properties.[30, 32, 33] First that the ordering of the positive operators can be interpreted as a hyperonomy relation of the words they represent (the ordering of positive operators being defined as : A ≤ B ⇔ B − A ≥ 0). Second, that when using positive operators for words, the compositions introduced in the DisCoCat framework are well defined and result in a positive operator as well.

When using this extension, the output of a circuit is no longer deterministic. By multiplying measurements of it, we can induce the positive operator for the represented sentence. The author further shows that the ordering of such positive operators representing sentences can be interpreted as an entailment relation.[30, 32, 33] Therefore, the author is able to use her framework to predict entailment of sentences.

## B. Experiments

In her experiments, the positive operators are built using GloVe vectors for elementary words and WordNet to derive hyponomy.[30, 32] The calculations are realized on a classical computer. The resulting model is then tested against BLESS, WBLESS and BIBLESS datasets for hyperonomy relations, and the KS2016 dataset for entailment. These datasets are made up of pairs of words (resp. sentences), that are labelled 0 or 1, depending on whether they are in a hyperonomy (resp. entailment) relation.

At the word level (Table 5.1), the model does not outperform best supervised model for hyperonomy prediction, although differences are minimal (0.01 accuracy). However, at the sentence level (Table 5.2), the model for entailment prediction outperforms previous best scores on the dataset.

By introducing an error term to the definition of ordering for positive operators, Martha Lewis further proposes an extension to her own model to account for *graded hyperonomy*. Hyperonomy is no longer a binary relation between two words, but is rather a weighted one. For instance, with this extension, "whale" could be modeled as only 70% hyponym of "mammal".

In M. Lewis, 2020, the author also provides an extension to her own model to account for negating words: by subtracting its positive operator to the identity operator (Fig. 5.4).[31, 32] This model is tested against KS2016 dataset (Table 5.3) and shows good results (93% average accuracy), thus showing compatibility with compositional hyperonomy.

| Model | BLESS | WBLESS | BIBLESS |
|---|---|---|---|
| HyperVec - WN | 0.92 | 0.87 | 0.81 |
| Hearst | 0.96 | 0.87 | 0.85 |
| HypeCones | 0.94 | 0.90 | 0.87 |
| LEAR - WN | 0.96 | 0.92 | 0.88 |
| Symb - WN | 0.91 | 0.93 | 0.91 |
| $k_{BA}$ - WN | 0.95 | 0.88 | 0.84 |
| $k_E$ - WN | 0.95 | 0.91 | 0.87 |
| $k_{BA}$ - Hearst | 0.91 | 0.84 | 0.76 |
| $k_E$ - Hearst | 0.91 | 0.86 | 0.80 |

Table 5.1 (M. Lewis [QNLP], 2019): comparison of accuracy of various models for hyperonomy prediction against three datasets (BLESS, WBLESS and BIBLESS). The four first models are the supervised approaches, the two following are symbolic models and the three last ones are the models using positive operators.

| | $k_E$ measure | | | $k_{BA}$ measure | | |
|---|---|---|---|---|---|---|
| Model | SV | VO | SVO | SV | VO | SVO |
| KS2016 best | 0.84 | 0.82 | 0.86 | 0.84 | 0.82 | 0.86 |
| Verb only | 0.632 | 0.632 | 0.663 | 0.868* | 0.829* | 0.890* |
| Addition | 0.576 | 0.586 | 0.492 | 0.893* | 0.892* | 0.945* |
| Mult | 0.885* | 0.842* | 0.966* | 0.961* | 0.934* | 0.980* |
| BMult1 | 0.794 | 0.749 | 0.880* | 0.945* | 0.916* | 0.977* |
| BMult2 | 0.778 | 0.723 | 0.869 | 0.949* | 0.914* | 0.980*+ |
| KMult1 | 0.881* | 0.833* | 0.946* | 0.957*+ | 0.934*+ | 0.984*+ |
| KMult2 | 0.823 | 0.800 | 0.930* | 0.909* | 0.939*+ | 0.963* |

Table 5.2 (M. Lewis [QNLP], 2019): comparison of accuracy of various models for entailment prediction against KS2016 dataset (using sentences of type SV, VO and SVO). The three first models are the supervised approaches, the five following are models using positive operators.

$$[\![not\ w]\!] := \mathbb{I} - [\![w]\!]$$

Fig. 5.4 (M. Lewis, 2020) : proposed formula to encode negation of words within paradigm of encoding words into positive operators

| Model | noun-verb | ¬noun-verb | noun-¬verb | ¬noun-¬verb |
|---|---|---|---|---|
| KS2016 best | 0.84 | - | - | - |
| Verb only | 0.866 | 0.867 | 0.865 | 0.867 |
| Noun only | 0.926 | 0.921 | 0.925 | 0.923 |
| Average | 0.947$^+$ | **0.946$^+$** | **0.948$^+$** | 0.946$^+$ |
| Mult | **0.960$^{*+}$** | 0.874 | 0.931$^+$ | **0.950$^+$** |
| BMult | 0.948$^+$ | 0.892 | 0.928 | 0.947$^+$ |
| BMult switched | 0.949$^+$ | 0.896 | 0.916 | 0.944$^+$ |
| KMult | 0.950$^+$ | 0.875 | 0.925 | 0.948$^+$ |
| KMult switched | 0.950$^+$ | 0.874 | 0.920 | 0.948$^+$ |

Table 5.3 (M. Lewis [QNLP], 2019) : comparison of accuracy of various models for entailment predictions against KS2016 dataset (using various sentences involving negations of the noun or the verbs). The three first models are the supervised approaches the five following are models using positive operators.

# VI – Conclusions

Throughout this work we have examined the DisCoCat framework, which is used to encode into a quantum circuit the meaning of a sentence (derived from the words' embeddings and the underlying syntactic structure). In particular, we have seen its applications through three use-cases: sentence similarity, question-answering and entailment prediction.

For the task of question-answering, we have presented our original experiments implementing logical connectors. We have done so in order to test the DisCoCat model against logical compositions in which the truth value of the composed sentence should be directly predictable from the truth value of the elements. We have discussed some concerns as to the scalability and the universality of the model used for the task, but have seen that these have been partly addressed in other experiments by scaling-up the circuits. While this raises valid further preoccupations at a time where the size of quantum computers are limited, we have proposed a promising implementation of logical connectors in the DisCoCat framework.

Our discussion has been a gentle survey of the literature, where we have introduced original experiments. We have first seen how the DisCoCat framework models the meaning of a sentence by combining the vectors representing its words' meanings along its syntactic structure. We then went through a DisCoCat inspired hybrid classical-quantum workflow to calculate sentence similarity. We further explored how to use DisCoCat to achieve question-answering. This method has shown improved understandability in comparison to classical alternatives. Finally we have presented a possible extension to the DisCoCat's model that can capture word hyperonomy and sentence entailment.

As we have seen throughout this work, the area of QNLP is still mainly at stage of proofs of concepts. Only one of the three use case we discussed has ever run on a real quantum computer. There is a need to scale to larger problems and to apply these methods to real industry cases. Generally, QNLP faces the same limitations as the rest of the quantum computing field: unrealized quantum RAM, limitation in the number of qubits, absence of fault tolerant universal quantum machine etc., which lead recent researches to focus on noisy intermediate-scale quantum applications. QNLP also faces specific challenges such as the implementation of logical operators, in particular *conversational logical operators*. Indeed logical operators have more subtle meaning than their mathematical counterparts : "Bob is not not rich" doesn't mean he is rich. Various recent works have focused on logical operators.[31, 44, 47]

We are conscious that as such, our experiments are merely proofs of concept. Just as the other experiments on question answering did, we could try increas-ing the size of our circuits to improve the scores. Still following what has been done previously, we could run the experiments on an actual quantum computer instead of a quantum simulator. This is particularly true now that the team from Cambridge Quantum Computing has released the brand new lambeq python library which allows to encode DisCoCat instances directly on quantum devices.[19, 20] We should also provide a non trivial exploration of the space of the parameters by computing the gradients of the qubits outputed by the circuits to run a gradient descent. Finally, we could explore more subtle implementations of logical connectors in order to account for their conversational meaning.